\documentclass[conference,letterpaper]{IEEEtran}

\IEEEoverridecommandlockouts    

\usepackage[utf8]{inputenc} 
\usepackage[T1]{fontenc}    
\usepackage{url}            
\usepackage{booktabs}       

\usepackage{microtype}      

\usepackage[cmex10]{amsmath} 
\usepackage{amsfonts,latexsym,amsthm,amssymb,amsmath,amscd,euscript}

\usepackage{bbm}
\usepackage{framed}
\usepackage{mathrsfs}

\usepackage{mathtools}
\usepackage{stmaryrd}

\usepackage[noadjust]{cite}

\usepackage{algorithm}
\makeatletter
\renewcommand{\ALG@name}{Algorithm}
\makeatother
\usepackage{algpseudocode}
\algblockdefx{ForAllP}{EndFaP}[1]%
{\textbf{for all }#1 \textbf{do in parallel}}%
{\textbf{end for}}
\algnewcommand\algorithmicforeach{\textbf{for each}}
\algdef{S}[FOR]{ForEach}[1]{\algorithmicforeach\ #1\ \algorithmicdo}

\usepackage{hyperref}
\hypersetup{
		colorlinks=true,
		linkcolor=black,
		linktoc=all,     
		filecolor=blue,
		citecolor = blue,      
		urlcolor=cyan,
		pdfpagemode=UseOutlines
	}
\usepackage{accents}
\usepackage{setspace}
	\usepackage[usenames,dvipsnames]{xcolor}
	\definecolor{plum}  {rgb}{.4,0,.4}
	\definecolor{bred} {rgb}{0.6,0,0}
	\definecolor{lnkcolr}{rgb}{0.55, 0.3, 0.09} 
	\definecolor{urlcolr}{rgb}{0.0, 0.35, 0.26}

	\usepackage{tikz-cd}
	\usepackage{tikz,pgfplots,pgfplotstable}
	\usetikzlibrary{matrix}
	\usetikzlibrary{cd}
	\usetikzlibrary{calc}
	\usetikzlibrary{arrows}      
	\usetikzlibrary{decorations.markings}
	\usetikzlibrary{positioning}
	\pgfplotsset{width=7cm,compat=1.8}  
	
	\usepackage{float}

	\usepackage{bm}

	\usepackage{nicefrac}
	\usepackage{enumitem}
	\usepackage{verbatim}
	\usepackage[bottom]{footmisc}
	\usepackage{tcolorbox}
	\usepackage[normalem]{ulem}
	\usepackage{nicematrix}
	\usepackage{adjustbox}
	\usepackage{changepage}
	\usepackage{multirow}
	\usepackage{wrapfig}
	
	\makeatletter
	\newtheorem*{rep@theorem}{\rep@title}
	\newcommand{\newreptheorem}[2]{%
		\newenvironment{rep#1}[1]{%
			\def\rep@title{#2 \ref{##1}}%
			\begin{rep@theorem}}%
			{\end{rep@theorem}}}
	\makeatother

	\theoremstyle{plain}
	\newtheorem{theorem}{Theorem}

	\newtheorem{definition}[theorem]{Definition}

	\theoremstyle{definition}
	\newtheorem{remark}[theorem]{Remark}
	\newtheorem{example}[theorem]{Example}
	
	\newtheorem*{tldr*}{TL;DR}
	\newtheorem*{impressions*}{Impressions}

	\newcommand{\ex}[2]{{\ifx&#1& \mathbb{E} \else \underset{#1}{\mathbb{E}} \fi \left[#2\right]}}
	\newcommand{\var}[2]{{\ifx&#1& \mathsf{Var} \else \underset{#1}{\mathsf{Var}} \fi \left[#2\right]}}
	
	\newcommand{\Rb}{\mathbb{R}}
	
	\usepackage{bbm}
	
	\def\1{\mathbf{1}}

	\let\originalleft\left
	\let\originalright\right
	\renewcommand{\left}{\mathopen{}\mathclose\bgroup\originalleft}
	\renewcommand{\right}{\aftergroup\egroup\originalright}

\newcommand{\mybignote}[2]{}

\definecolor{bblue}{rgb}{0.0, 0.18, 0.39}
\definecolor{bluee}{rgb}{0.33, 0.41, 0.92}
\definecolor{redd}{rgb}{0.99, 0.4, 0.37}
\definecolor{blblue}{rgb}{0.204, 0.482, 0.678}
\definecolor{darkgray}{rgb}{0.66, 0.66, 0.66}

\newcommand{\CMI}[2]{{\ifx&#2& \mathsf{CMI} \else \mathsf{CMI}_{#2} \fi \left(#1\right)}}

\usepackage{etoolbox}
\newcommand{\define}[4]{\expandafter#1\csname#3#4\endcsname{#2{#4}}}
\forcsvlist{\define{\newcommand}{\mathcal}{c}}{K,C,F,H,S,X,Y,Z,E,D,N,L,G,P,Q,A,T,I,M,W,B,U,V,O,R}  

\usepackage{thmtools}

\graphicspath{{./figures/}}

\title{Oversquashing in GNNs through the lens of \\ information contraction and graph expansion}

\author{Pradeep Kr.~Banerjee$^\ast$, Kedar Karhadkar$^\dagger$, Yu Guang~Wang$^\ddagger$, Uri Alon$^\S$, Guido Mont\'ufar$^\dagger$$^\ast$
	\thanks{$^{\ast}$MPI MiS, $^\dagger$UCLA, $^\ddagger$SJTU, $^\S$CMU}
	\thanks{{$^\ast$Correspondence to: \tt\small{pradeep@mis.mpg.de}}}
}

\begin{document}

	\maketitle

	\begin{abstract}
		The quality of signal propagation in message-passing graph neural networks (GNNs) strongly influences their expressivity as has been observed in recent works. In particular, for prediction tasks relying on long-range interactions, recursive aggregation of node features can lead to an undesired phenomenon called ``oversquashing''. We present a framework for analyzing oversquashing based on information contraction. Our analysis is guided by a model of reliable computation due to von Neumann that lends a new insight into oversquashing as signal quenching in noisy computation graphs. Building on this, we propose a graph rewiring algorithm aimed at alleviating oversquashing. Our algorithm employs a random local edge flip primitive motivated by an expander graph construction. We compare the spectral expansion properties of our algorithm with that of an existing curvature-based non-local rewiring strategy. Synthetic experiments show that while our algorithm in general has a slower rate of expansion, it is overall computationally cheaper, preserves the node degrees exactly and never disconnects the graph. 
		%
		%
	\end{abstract}

	\section{Introduction}
	
	Graph neural networks (GNNs) provide a powerful framework \cite{scarselli2008graph,kipfwelling} for modeling complex structural and relational data from diverse domains ranging from biological and social networks to knowledge graphs and molecules \cite{wu2020comprehensive}. The input to a GNN is a graph endowed with node embeddings or features. Its outputs are node representations that depend on both the structure of the graph and its features. GNNs broadly follow a message-passing paradigm \cite{gilmer2017neural,dai2016discriminative}, where feature representations for each node are learned by recursively aggregating and transforming representations of its neighbors. The range of aggregation is given by the number of GNN layers. Aggregation functions are locally permutation-invariant and come in different flavors and their specific form distinguishes different GNN variants \cite{kipfwelling,velivckovic2017graph,hamilton2017inductive,xu2018powerful}. Empirically, several of these variants have achieved state-of-the-art performance in tasks related to node and graph classification, clustering and link prediction. 
	
	Despite their empirical successes, the expressivity of GNNs is limited particularly for tasks involving long-range node interactions \cite{alon2021on,dwivedi2022long}. For example, a molecular property might depend on a pair of atoms residing on opposite ends of the molecule. 	Clearly, for one atom to be ``aware'' of its distant pair, it must draw information from nodes that are $r$ hops away, where $r$ is the \emph{problem radius} or range of the interaction, which in this extreme case is the molecule diameter. Learning such tasks is markedly complex and requires a large number of GNN layers. An $L$-layer GNN can capture structural information within at most an $L$-hop neighborhood of each node. So, for the example at hand, $L$ must at least be $r$ to capture the molecular interaction. For long-range tasks such as these, increasing the number of layers, however, comes at a cost and can potentially lead to a phenomenon called ``oversquashing'' \cite{alon2021on}. 
	
	Oversquashing arises from the recursive nature of the neighborhood aggregation process. For an $L$-layer GNN, ``unrolling'' this process results in a collection of subtrees (or {computation graphs}) of depth $L$ with each subtree rooted at one of the nodes of the graph \cite{xu2018representation,garg2020generalization}. Specifically, the subtree rooted at some node $v$ represents the $L$-hop neighborhood of $v$, where the children of any node $u$ in the tree are the nodes adjacent to $u$. See Figure \ref{fig:noisyBooleanCkt}(b) for an illustration. Information flows all the way up from the leaves to the root: Each leaf is endowed with a feature vector and the representation of $v$ is computed bottom-up by recursively aggregating the feature vectors from the subtrees. For prediction tasks with a problem radius of $r$, the depth $L$ of each local subtree must at least be $r$ to support the flow the information. With increasing $L$, however, information from exponentially-growing neighborhoods need to be concurrently propagated at each message-passing step. This leads to a bottleneck that causes oversquashing, when an exponential amount of information is squashed into fixed-size node vectors \cite{alon2021on}. As a result, the information content of signals flowing along the ensuing ``noisy'' computation graphs decays and the GNN fails to fit long-range patterns. 
	
	In this work, we seek a better conceptual understanding of oversquashing using tools from information theory and expander graph theory. Broadly, the content is divided into two parts: In the first part we present an information-theoretic framework for {analyzing} the information decay arising from oversquashing. In the second part, we propose an expander-based graph rewiring strategy aimed at {alleviating} oversquashing. 
	Our approach is guided by two interrelated themes: 
	\begin{itemize}[leftmargin=*]
		\item 
		{\bf Information bottleneck in noisy computation graphs:} 
		In information theory, mutual information contraction along Markov chains is quantitatively captured by the well-known data processing inequalities \cite{polyanskiybook}. A strong version of these inequalities can be obtained by introducing appropriate channel-dependent ``contraction coefficients'' that capture the rate of information loss when a signal propagates through a noisy channel \cite{dawsonevansshulman1999,raginsky2016strong,polyanskiy2017strong}. In the context of oversquashing, we show how the signal decay in noisy computation graphs is conceptually analogous to information contraction in a noisy circuit model due to von Neumann \cite{von1956probabilistic}. 
		
		\item {\bf Structural bottlenecks and expander graphs:}
		Since traditional GNNs use the input graph to propagate neural messages \cite{gilmer2017neural}, structural characteristics of the input graph play a crucial role in the quality of signal propagation across nodes. Expanders are graphs that have high isoperimetry: Every part of the graph is connected to the rest of it by a large fraction of its edges \cite{hoory2006expander}. In this sense, expander graphs have \emph{no} structural bottlenecks. They also have several nice properties, e.g., logarithmic diameter, rapid mixing of diffusions and random walks, etc.\ \cite{hoory2006expander,sarnak2004expander}. These graphs were originally studied for designing robust telephone networks \cite{pinsker1973complexity} with the goal of maintaining good connectivity even when some nodes fail. Since then, these graphs have found myriad applications \cite{hoory2006expander}, such as design of communication networks and error correction codes, to name a few. There is also some speculation that the brain as a graph is a good expander \cite{kolmogorov1967realization,valiant2014must}. Our graph rewiring strategy for addressing oversquashing is inspired by an expander graph construction. 
	\end{itemize}

	\textbf{Related work}
	Graph rewiring strategies for improving the expressivity of GNNs seek to 	decouple the input graph from its computational graph. Rewiring can assume different forms ranging from neighbourhood sampling \cite{hamilton2017inductive} and connectivity diffusion \cite{klicpera2019diffusion} to virtual nodes \cite{battaglia2018relational} and edge \cite{Rong2020DropEdge} or node dropout \cite{papp2021dropgnn}. 	Rewiring strategies that exclusively address the oversquashing problem are the +FA method \cite{alon2021on} and the Stochastic Discrete Ricci Flow (SDRF) method \cite{topping2022understanding}. In the +FA method \cite{alon2021on}, the last GNN layer is made an expander -- the complete graph that allows every pair of nodes to connect to each other and pass long-range signals that might otherwise get squashed. SDRF \cite{topping2022understanding} is a curvature-based non-local rewiring strategy that aims at alleviating structural bottlenecks by adding ``supporting'' edges around negatively curved edges while preserving the node degree distribution.

	\textbf{Contributions} 
	We summarize our main contributions: 
	\begin{itemize}[leftmargin=*]
		\item 
		We present a framework for analyzing the information decay arising from oversquashing in GNNs and show how an information percolation bound (Theorem~\ref{prop:EWS99MIbound}) captures the essence of oversquashing as signal quenching in noisy computation graphs. 
		\item 
		We propose a new {local} graph rewiring algorithm, the Greedy Random Local Edge Flip (G-RLEF) motivated by an expander graph construction. We compare the rewiring dynamics of G-RLEF with that of SDRF \cite{topping2022understanding}. Synthetic experiments show that while the SDRF can in general have a faster rate of expansion, our G-RLEF preserves the node degrees exactly and never disconnects the graph. 
	\end{itemize}

	\section{Preliminaries}\label{sec:prelim}
	We collect some basic notations and definitions. 
	Let $G = (\cV,\cE)$ be a simple undirected graph on $n$ nodes and $m$ edges. For two distinct nodes $u,v\in \cV$, the graph distance $d_G(u, v)$ is the number of edges in the shortest path connecting $u$ and $v$. The diameter of $G$ is $\max_{u,v} d_G(u, v)$. The {neighborhood} $N_G(u)$ of a node $u\in \cV$ is the set $\{v \colon \left(u, v\right) \in \cE\}$. The {degree} of $u \in \cV$ is the size of its neighborhood $|N_G(u)|$. The Laplacian of $G$ is $L=D-A$, where $D$ is the degree matrix and $A$ is the adjacency matrix. 
	If $G$ is connected, then $L$ has rank $n-1$, with its kernel spanned by the vector of all $1$’s. 
	
	\textbf{Graph neural networks} 
	Traditional graph neural networks follow a message-passing paradigm \cite{gilmer2017neural}, where each message-passing step is parameterized by a neural network layer. Nodes have embeddings at each layer. The embedding of a node $v$ in Layer-0 is its input feature vector. The embedding in Layer-$l$ is obtained by aggregating for every $v\in \cV$ simultaneously the neighborhood feature vectors from the previous layer using some parametric function $f_l$: 
	\begin{align}\label{eq:MPNNupate}
		{h}_{v}^{(l)}=f_{l}\left({h}_{v}^{(l-1)},\{{h}_{u}^{(l-1)} \colon u \in N_{G}(v)\right).
	\end{align}
	In an $L$-layer GNN, the final embedding $h_v^L$ of node $v$ gets structural information from nodes that are at most $L$ hops away. By unrolling the aggregation steps starting at node $v$, we obtain $v$'s \emph{computation graph}, a tree of depth $L$ rooted at $v$ that represents the $L$-hop neighborhood of $v$, where the children of any node $u$ in the tree are the nodes adjacent to $u$. 
	
	\section{Oversquashing and information decay \\in noisy computation graphs} 
	\label{sec:SigPropNoisyCkt}
	\subsection{Information contraction in Markov chains}
	To set the stage we first recall the data processing inequality (DPI), which states that {the mutual information satisfies} $I(U ; Y) \leq I(U ; X)$ for any Markov chain $U \rightarrow X \rightarrow Y$, i.e., we cannot gain information under the action of a noisy channel. In many cases where we strictly lose information it is possible to show that $I(U ; Y) < I(U ; X)$. This is captured by a quantitative version of the DPI called the \emph{strong data processing inequality} (SDPI) \cite{raginsky2016strong,polyanskiy2017strong}. We denote by $K$ the channel $X \rightarrow Y$ with finite input and output alphabets resp. $\cX$ and $\cY$, and transition probabilities $\{K(x,y) : x \in \cX,y \in \cY\}$. We denote by $\mathbb{P}_{\cX}$ the set of all probability measures on a finite set $\cX$. We say that the channel $K$ satisfies a SDPI if $I(U ; Y) \leq \eta_{\mathrm{KL}}(K) I(U ; X)$, where $\eta_{\mathrm{KL}} \in [0,1]$ is called the \emph{Kullback-Leibler (KL) contraction coefficient} of the channel, which is defined as  
	\begin{align}
		\eta_{\mathrm{KL}}(K) :=\sup_{\substack{P,Q \in \mathbb{P}_{\cX}: \\ 0 < D(P\|Q) < \infty}} \frac{D(K\circ P \| K\circ Q)}{D(P \| Q)},
	\end{align}
	where $K\circ P$ is the distribution on $\cY$ induced by the push-forward of $P\in\mathbb{P}_\cX$ and $D(P\|Q):=\int \log \frac{dP}{dQ} dP$ is the KL divergence. The KL contraction coefficient admits the following alternative characterization in terms of the contraction of mutual information \cite{anantharamSDPI}:
	\begin{align}
		\eta_{\mathrm{KL}}(K) &=\sup _{U \rightarrow X \rightarrow Y} \frac{I(U ; Y)}{I(U ; X)} . 
	\end{align}
	\begin{example}
		Let $K$ is the binary symmetric channel (BSC) with crossover probability $\delta \in (0,\tfrac{1}{2})$, denoted by $\textsf{\small BSC}(\delta)$, i.e., $Y= X \oplus Z :=X+Z \mod 2$, where $Z\sim \rm{Bernoulli}(\delta)$ is independent of $X$.
		When information passes through a $\textsf{\small BSC}(\delta)$, it gets lost by a fraction $\eta_{\mathrm{KL}}(\textsf{\small BSC}(\delta))=(1-2\delta)^2$ \cite{ahlswede1976spreading}.
	\end{example}
	
	\begin{remark}[KL contraction coefficient and the Ollivier-Ricci curvature]\label{rem:etaKLCurvature}
		Given a locally finite, simple, connected undirected graph $G=(\cV,\cE)$, let $K_G=D^{-1}A$ be the channel associated with a simple random walk over the nodes of $G$. Define the \emph{Kantorovich norm} of $K_G$ by 
		\begin{align}
			\tau(K_G):= \sup _{x, x' \in \cV, x \neq x'} \frac{W_1(K_G(x,\cdot), K_G(x',\cdot))}{d_G(x, x')},
		\end{align}
		where $W_1(\cdot,\cdot)$ is the 1-Wasserstein distance on $\mathbb{P}_{\mathcal{V}}$ with ground metric the graph distance $d_G$. The Kantorovich norm was introduced by Dobrushin \cite{dobrushin1956centralI,dobrushin1996perturbation} and is also called the \emph{generalized Dobrushin's ergodicity coefficient} 	\cite{del2003contraction}. The \emph{Ollivier-Ricci curvature} \cite{ollivier2009ricci} of $G$ is then 
		\begin{align}\label{eq:ORcurvaturegraphs}
			\kappa(G) := 1-\tau(K_G).
		\end{align}
		Nonnegative Ollivier-Ricci curvature is thus equivalent to the requirement that the channel $K_G$ is a contraction under the 1-Wasserstein metric. 
		For any channel $K$, we have $\eta_{\mathrm{KL}}(K) \le  \tau(K)$ \cite{CohenKempermanZbaganu98}, so that the KL contraction coefficient of $K_G$ is always upper-bounded by $1-\kappa(G)$ when the Ollivier-Ricci curvature is nonnegative.
	\end{remark}
	\subsection{Reliable signal propagation in noisy Boolean circuits}\label{sec:noisycktmodel}
\begin{figure}
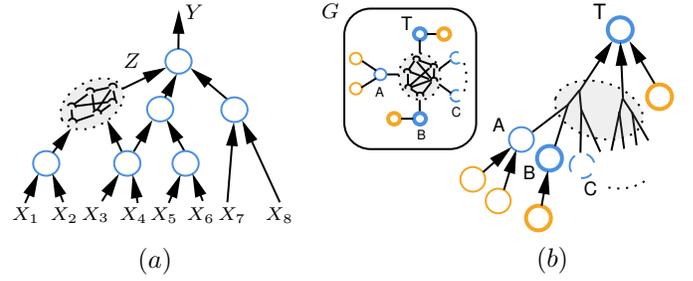

		\centering
		
		\tikzset{every picture/.style={line width=0.75pt}} 
		


		\caption{(a) A 8-input Boolean circuit comprising of $\delta$-noisy gates with fan-in at most $k=3$. $Z$ is the output of a computation involving a long chain of $\delta$-noisy gates (shaded in grey). 
			(b) The computation graph of the target node \textsf{T}, and input graph $G$ (inset) for the \textsc{NeighborsMatch} problem (see text). The shaded region is an unknown subgraph. 
			For predicting the label of the root \textsf{T}, the network must concurrently propagate information bottom-up from all the leaves to the root, and make a decision given a single fixed-sized node vector residing at \textsf{T} that compresses all this information.
		}
		\label{fig:noisyBooleanCkt}
	\end{figure}

	A \emph{Boolean circuit} with $n$ inputs is a directed acyclic graph (DAG) in which nodes of in-degree zero are either references to the inputs $(X_1,\ldots,X_n)$ or Boolean constants (0 or 1) and nodes of in-degree at most $k\ge 1$ are logic gates computing Boolean functions of at most $k$ arguments. The output of the circuit is a unique node of out-degree zero. A \emph{noisy Boolean circuit} is composed of Boolean gates that fail (i.e., produce a $0$ instead of a $1$ or vice versa) independently with probability $\delta \le 1/2$. We refer to these gates as \emph{$\delta$-noisy gates}. Figure~\ref{fig:noisyBooleanCkt}(a) shows an example of a 8-input Boolean circuit comprising of {$\delta$-noisy gates} with in-degree or fan-in at most $k=3$.
	
	Von Neumann \cite{von1956probabilistic} asked the following question: 
	Can every Boolean circuit with noiseless gates be simulated by a noisy Boolean circuit? He showed that if $\delta$ is sufficiently small, then there exists $\epsilon > 1/2$ such that for any Boolean function $f\colon \{0,1\}^n\to \{0,1\}$ there is a noisy circuit $C$ that correctly computes $f$ on every input with probability at least $\epsilon$. Von Neumann's noisy circuit model was roughly inspired by McCulloch and Pitts neural network model \cite{mccullochpitts1943}, only now with the adjunction of a finite probability space to model noisy gates \cite{pippenger1990developments}. 	Later Pippenger \cite{pippenger1985networks} gave an explicit construction of $C$ that entails using 3-majority gates as ``expanders'' for local error correction \cite{wigdersonexpandertalk}. 	
	
	For $C$ to correctly compute $f$ for every input with probability better than random guessing, the mutual information between the inputs and the output of the noisy circuit should be positive. This mutual information is constrained by the structure of the intervening noisy circuit. For a large enough circuit,	the output will have little correlation with most of the inputs if the gates are too noisy \cite{dawsonevansshulman1999}. This might happen, for example, for the left-most branch in the circuit in Figure~\ref{fig:noisyBooleanCkt}(a) where the computation involves a very long chain of $\delta$-noisy gates. The following theorem due to Evans and Schulman \cite{dawsonevansshulman1999} gives an upper bound on the input-output mutual information:
	\begin{theorem}\label{prop:EWS99MIbound}
		Consider an $n$-input noisy Boolean circuit composed of gates with fan-in at most $k$ where each gate fails (produces a 0 instead of a 1 or vice versa) independently with probability at most $\delta$. Then, the mutual information between any input $X_i$ and output $Y$ is upper bounded as
		\begin{align}\label{eq:ES-MI-bound}
			I(X_i;Y) \le (\eta \cdot k)^{d_i}\log 2,
		\end{align}
		where $\eta = \eta_{\mathrm{KL}}(\textsf{\small BSC}(\delta))=(1-2\delta)^2$, and $d_i$ is the graph distance between $X_i$ and $Y$.
	\end{theorem}
	Theorem~\ref{prop:EWS99MIbound} implies that reliable computation is possible only when $\delta < \tfrac{1}{2}-\tfrac{1}{2\sqrt{k}}$. Polyanskiy and Wu \cite{polyanskiy2017strong} obtained improved upper bounds by relating $\eta_{\mathrm{KL}}$ to the probability of site percolation on the DAG. When the underlying graph is a tree, for $k$ odd and large, reliable computation is possible if and only if $\delta < \tfrac{1}{2}-\tfrac{\sqrt{\pi/2}}{2\sqrt{k}}$ \cite{dawsonevansshulman2003} showing the tightness of Theorem~\ref{prop:EWS99MIbound}.

	\subsection{Information contraction in noisy computation graphs from oversquashing} \label{sec:neighborsmatch}
	For problems featuring a long-range dependence between nodes in the circuit DAG (see Figure~\ref{fig:noisyBooleanCkt}(a)), small values of the input-output mutual information are indicative of high information contraction. 
	This is analogous to the situation arising from oversquashing. 
	Consider the \textsc{NeighborsMatch} problem introduced in \cite{alon2021on}. 
	Given an input graph $G$ (see inset in Figure~\ref{fig:noisyBooleanCkt}(b)), suppose that we wish to predict the label for 
	a node \textsf{T}. 
	The correct label is the label of the blue node that has the same number of orange neighbors as 
	\textsf{T}. 
	Each example in the training dataset is a different input graph with a different mapping from numbers of neighbors to labels.
	For the example in Figure~\ref{fig:noisyBooleanCkt}(b), the answer is \textsf{B} 
	which happens to reside at the opposite end of the 
	graph. 
	Thus, correctly predicting the label for this example will require a number of GNN layers $L$ that is equal to the diameter of the graph to capture the long-range dependence between \textsf{T} and \textsf{B}. 
	With increasing $L$, however, at each message-passing layer, information from exponentially-growing neighborhoods need to be concurrently propagated; 
	see Figure~\ref{fig:noisyBooleanCkt}(b).
	%
	This leads to oversquashing 
	when the nodes in the computation graph behave like ``noisy gates'' and the GNN fails to fit the training dataset perfectly.
	For the \textsc{NeighborsMatch} problem, oversquashing can start affecting 
	some GNNs even for $L=4$ layers, 
	and increasing the dimension of the hidden node feature vectors for a given problem radius leads only to a marginal improvement in training accuracy;
	see \cite[Figures 3 and 4]{alon2021on}.

	%
	The relevance of the information contraction argument (Theorem~\ref{prop:EWS99MIbound}) in this context is evident from noticing that 
	if $\eta < 1/k$, then we have that 
	$(\eta \cdot k)^{d_i} \to 0$ 
	and the mutual information $I(X_i;Y)$ vanishes as the graph distance $d_i$ between $X_i$ and $Y$ increases. 
	By way of a rough analogy, for the \textsc{NeighborsMatch} problem, $d_i$ will correspond to the number of layers $L$ in the computation graph of the target node \textsf{T} and $k$ to the maximum degree of the nodes in $G$ in Figure~\ref{fig:noisyBooleanCkt}(b). 
	Given an input graph, the maximum degree $k$ is fixed. 
	Then the condition $\eta < 1/k$ can be satisfied 
	when an exponential amount of information is squashed into a fixed-size node vector leading to oversquashing, which translates into a high rate of information loss or low $\eta$.

	
	\section{Structural bottlenecks and \\ Expander-based graph rewiring algorithms}
	\label{sec:graphrewire}
	
	\subsection{Spectral gap and graph expansion}
	\label{sec:spectralgap-expansion}
	
	We briefly review some facts about expander graphs \cite{hoory2006expander,sarnak2004expander}. For simplicity of exposition, we restrict our attention to unweighted regular graphs. 
	
	Let $G$ be a $d$-regular graph on $n$ nodes. We number the eigenvalues of the adjacency matrix $A$ in decreasing order: $d=\mu_1\ge \mu_2\ge \cdots \ge \mu_n$. The difference $d -\mu_2(A)$ is referred to as the \emph{spectral gap} of $G$. We write $\mu(A) = \max(|\mu_2|, |\mu_n|)$ to denote the largest absolute eigenvalue of $A$ other than $\mu_1 = d$.
	\begin{definition} 
		Given two sets $\cS,\cT\subset \cV$, the set of edges between $\cS$ and $\cT$ is denoted $E(\cS,\cT)=\{(u,v)\colon u \in \cS, v\in \cT, (u,v)\in \cE\}$.
		The \emph{edge boundary} of a set $\cS\subset \cV$, denoted $\partial \cS$, is 
		$\partial \cS =E(\cS,\bar{\cS})$, where $\bar{\cS}=\cV\setminus \cS$. 
		The \emph{isoperimetric ratio} or the \emph{Cheeger constant} of $G$ is
		\begin{align}\label{eq:isoperimetricratio}
			h(G)=\min_{\cS\subset \cV\colon |\cS|\le n/2} \frac{|\partial \cS|}{|\cS|}.
		\end{align}
		A $d$-regular graph $G$ on $n$ nodes is a \emph{$(n,d,\beta)$-expander} if $h(G)\ge \beta$, where $\beta$ is a constant independent of $n$. An infinite family $(G_i)_{i\ge 1}$ of $(n_i,d,\beta)$-expanders 		forms an expander family if $h(G_i)\ge \beta$ for all $i$. 
	\end{definition}
	The isoperimetric ratio $h(G)$ is large if and only if every subset of at most half the nodes has a large surface area to volume ratio. Of course, all finite connected graphs are expanders for some $\beta>0$ and theoretically, the case $d=O(1)$ and $n\gg d$ for sparse, large expanders is the most well-studied for their extremal properties.

	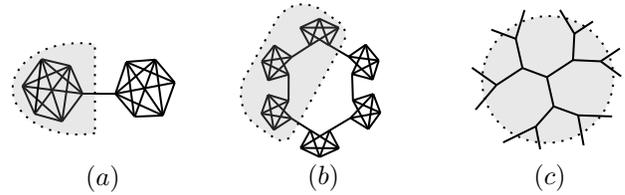
\begin{figure}
		\centering
		
		\tikzset{every picture/.style={line width=0.75pt}} 
		
		\begin{tikzpicture}[x=0.75pt,y=0.75pt,yscale=-1,xscale=1]
			
			\draw [line width=0.75]    (187.36,11607.11) -- (197.06,11602.63) ;
			\draw [line width=0.75]    (197.06,11602.63) -- (206.76,11607.11) ;
			\draw [line width=0.75]    (201.91,11616.06) -- (206.76,11607.11) ;
			\draw [line width=0.75]    (192.21,11616.06) -- (187.36,11607.11) ;
			\draw [line width=0.75]    (192.21,11616.06) -- (206.76,11607.11) ;
			\draw [line width=0.75]    (187.36,11607.11) -- (201.91,11616.06) ;
			\draw [line width=0.75]    (187.36,11607.11) -- (206.76,11607.11) ;
			\draw [line width=0.75]    (192.21,11616.06) -- (197.06,11602.63) ;
			\draw [line width=0.75]    (197.06,11602.63) -- (201.91,11616.06) ;
			\draw [line width=0.75]    (183.11,11621.93) -- (192.21,11616.06) ;
			\draw [line width=0.75]    (171.17,11633.72) -- (167.96,11624.16) ;
			\draw [line width=0.75]    (167.96,11624.16) -- (174.32,11616.06) ;
			\draw [line width=0.75]    (183.11,11621.93) -- (174.32,11616.06) ;
			\draw [line width=0.75]    (181.53,11630.76) -- (171.17,11633.72) ;
			\draw [line width=0.75]    (181.53,11630.76) -- (174.32,11616.06) ;
			\draw [line width=0.75]    (171.17,11633.72) -- (183.11,11621.93) ;
			\draw [line width=0.75]    (171.17,11633.72) -- (174.32,11616.06) ;
			\draw [line width=0.75]    (181.53,11630.76) -- (167.96,11624.16) ;
			\draw [line width=0.75]    (167.96,11624.16) -- (183.11,11621.93) ;
			
			\draw [line width=0.75]    (174.03,11658.69) -- (167.8,11650.5) ;
			\draw [line width=0.75]    (167.8,11650.5) -- (171.16,11640.98) ;
			\draw [line width=0.75]    (181.47,11644.08) -- (171.16,11640.98) ;
			\draw [line width=0.75]    (182.91,11652.93) -- (174.03,11658.69) ;
			\draw [line width=0.75]    (182.91,11652.93) -- (171.16,11640.98) ;
			\draw [line width=0.75]    (174.03,11658.69) -- (181.47,11644.08) ;
			\draw [line width=0.75]    (174.03,11658.69) -- (171.16,11640.98) ;
			\draw [line width=0.75]    (182.91,11652.93) -- (167.8,11650.5) ;
			\draw [line width=0.75]    (167.8,11650.5) -- (181.47,11644.08) ;
			
			\draw [line width=0.75]    (181.53,11630.76) -- (181.47,11644.08) ;
			\draw [line width=0.75]    (206.92,11668) -- (197.17,11672.38) ;
			\draw [line width=0.75]    (197.17,11672.38) -- (187.52,11667.82) ;
			\draw [line width=0.75]    (192.47,11658.91) -- (187.52,11667.82) ;
			\draw [line width=0.75]    (202.16,11659) -- (206.92,11668) ;
			\draw [line width=0.75]    (202.16,11659) -- (187.52,11667.82) ;
			\draw [line width=0.75]    (206.92,11668) -- (192.47,11658.91) ;
			\draw [line width=0.75]    (206.92,11668) -- (187.52,11667.82) ;
			\draw [line width=0.75]    (202.16,11659) -- (197.17,11672.38) ;
			\draw [line width=0.75]    (197.17,11672.38) -- (192.47,11658.91) ;
			
			\draw [line width=0.75]    (182.91,11652.93) -- (192.39,11658.96) ;
			\draw [line width=0.75]    (212.7,11652.73) -- (202.16,11659) ;
			\draw [line width=0.75]    (224.19,11640.57) -- (227.75,11650.02) ;
			\draw [line width=0.75]    (227.75,11650.02) -- (221.7,11658.32) ;
			\draw [line width=0.75]    (212.7,11652.73) -- (221.7,11658.32) ;
			\draw [line width=0.75]    (213.95,11643.85) -- (224.19,11640.57) ;
			\draw [line width=0.75]    (213.95,11643.85) -- (221.7,11658.32) ;
			\draw [line width=0.75]    (224.19,11640.57) -- (212.7,11652.73) ;
			\draw [line width=0.75]    (224.19,11640.57) -- (221.7,11658.32) ;
			\draw [line width=0.75]    (213.95,11643.85) -- (227.75,11650.02) ;
			\draw [line width=0.75]    (227.75,11650.02) -- (212.7,11652.73) ;
			
			\draw [line width=0.75]    (220.41,11615.71) -- (226.95,11623.69) ;
			\draw [line width=0.75]    (226.95,11623.69) -- (223.94,11633.31) ;
			\draw [line width=0.75]    (213.52,11630.53) -- (223.94,11633.31) ;
			\draw [line width=0.75]    (211.76,11621.73) -- (220.41,11615.71) ;
			\draw [line width=0.75]    (211.76,11621.73) -- (223.94,11633.31) ;
			\draw [line width=0.75]    (220.41,11615.71) -- (213.52,11630.53) ;
			\draw [line width=0.75]    (220.41,11615.71) -- (223.94,11633.31) ;
			\draw [line width=0.75]    (211.76,11621.73) -- (226.95,11623.69) ;
			\draw [line width=0.75]    (226.95,11623.69) -- (213.52,11630.53) ;
			
			\draw [line width=0.75]    (213.95,11643.85) -- (213.52,11630.53) ;
			\draw [line width=0.75]    (211.76,11621.73) -- (202.06,11616.01) ;
			\draw  [fill={rgb, 255:red, 155; green, 155; blue, 155 }  ,fill opacity=0.23 ][dash pattern={on 0.84pt off 2.51pt}][line width=0.75]  (281.26,11622.5) .. controls (287.74,11605.86) and (306.86,11597.21) .. (323.95,11603.16) .. controls (341.05,11609.11) and (349.65,11627.42) .. (343.16,11644.05) .. controls (336.68,11660.69) and (317.56,11669.34) .. (300.47,11663.39) .. controls (283.37,11657.44) and (274.77,11639.13) .. (281.26,11622.5) -- cycle ;
			\draw [color={rgb, 255:red, 0; green, 0; blue, 0 }  ,draw opacity=1 ][fill={rgb, 255:red, 155; green, 155; blue, 155 }  ,fill opacity=0.23 ]   (299.44,11627.49) -- (311.8,11632.1) ;
			\draw [color={rgb, 255:red, 0; green, 0; blue, 0 }  ,draw opacity=1 ][fill={rgb, 255:red, 155; green, 155; blue, 155 }  ,fill opacity=0.23 ]   (311.8,11632.1) -- (315.17,11646.85) ;
			\draw [color={rgb, 255:red, 0; green, 0; blue, 0 }  ,draw opacity=1 ][fill={rgb, 255:red, 155; green, 155; blue, 155 }  ,fill opacity=0.23 ]   (299.44,11627.49) -- (286.32,11636.09) ;
			\draw [color={rgb, 255:red, 0; green, 0; blue, 0 }  ,draw opacity=1 ][fill={rgb, 255:red, 155; green, 155; blue, 155 }  ,fill opacity=0.23 ]   (296.08,11612.74) -- (299.44,11627.49) ;
			\draw [color={rgb, 255:red, 0; green, 0; blue, 0 }  ,draw opacity=1 ][fill={rgb, 255:red, 155; green, 155; blue, 155 }  ,fill opacity=0.23 ]   (324.93,11623.49) -- (311.8,11632.1) ;
			\draw [color={rgb, 255:red, 0; green, 0; blue, 0 }  ,draw opacity=1 ][fill={rgb, 255:red, 155; green, 155; blue, 155 }  ,fill opacity=0.23 ]   (315.17,11646.85) -- (327.57,11651.37) ;
			\draw [color={rgb, 255:red, 0; green, 0; blue, 0 }  ,draw opacity=1 ][fill={rgb, 255:red, 155; green, 155; blue, 155 }  ,fill opacity=0.23 ]   (324.93,11623.49) -- (338.89,11624.1) ;
			\draw [color={rgb, 255:red, 0; green, 0; blue, 0 }  ,draw opacity=1 ][fill={rgb, 255:red, 155; green, 155; blue, 155 }  ,fill opacity=0.23 ]   (306.22,11656.95) -- (315.17,11646.85) ;
			\draw [color={rgb, 255:red, 0; green, 0; blue, 0 }  ,draw opacity=1 ][fill={rgb, 255:red, 155; green, 155; blue, 155 }  ,fill opacity=0.23 ]   (324.93,11623.49) -- (325.62,11610.19) ;
			\draw [color={rgb, 255:red, 0; green, 0; blue, 0 }  ,draw opacity=1 ][fill={rgb, 255:red, 155; green, 155; blue, 155 }  ,fill opacity=0.23 ]   (275.8,11648.52) -- (286.32,11636.09) ;
			\draw [color={rgb, 255:red, 0; green, 0; blue, 0 }  ,draw opacity=1 ][fill={rgb, 255:red, 155; green, 155; blue, 155 }  ,fill opacity=0.23 ]   (296.08,11612.74) -- (300.48,11598.21) ;
			\draw [color={rgb, 255:red, 0; green, 0; blue, 0 }  ,draw opacity=1 ][fill={rgb, 255:red, 155; green, 155; blue, 155 }  ,fill opacity=0.23 ]   (286.32,11636.09) -- (273.97,11627.34) ;
			\draw [color={rgb, 255:red, 0; green, 0; blue, 0 }  ,draw opacity=1 ][fill={rgb, 255:red, 155; green, 155; blue, 155 }  ,fill opacity=0.23 ]   (283.72,11608.13) -- (296.08,11612.74) ;
			\draw [color={rgb, 255:red, 0; green, 0; blue, 0 }  ,draw opacity=1 ][fill={rgb, 255:red, 155; green, 155; blue, 155 }  ,fill opacity=0.23 ]   (325.62,11610.19) -- (318.38,11598.2) ;
			\draw [color={rgb, 255:red, 0; green, 0; blue, 0 }  ,draw opacity=1 ][fill={rgb, 255:red, 155; green, 155; blue, 155 }  ,fill opacity=0.23 ]   (308.39,11668.47) -- (306.16,11656.92) ;
			\draw [color={rgb, 255:red, 0; green, 0; blue, 0 }  ,draw opacity=1 ][fill={rgb, 255:red, 155; green, 155; blue, 155 }  ,fill opacity=0.23 ]   (330.29,11666.52) -- (327.51,11651.35) ;
			\draw [color={rgb, 255:red, 0; green, 0; blue, 0 }  ,draw opacity=1 ][fill={rgb, 255:red, 155; green, 155; blue, 155 }  ,fill opacity=0.23 ]   (349.21,11631.69) -- (338.89,11624.1) ;
			\draw [color={rgb, 255:red, 0; green, 0; blue, 0 }  ,draw opacity=1 ][fill={rgb, 255:red, 155; green, 155; blue, 155 }  ,fill opacity=0.23 ]   (325.62,11610.19) -- (335.01,11603.99) ;
			\draw [color={rgb, 255:red, 0; green, 0; blue, 0 }  ,draw opacity=1 ][fill={rgb, 255:red, 155; green, 155; blue, 155 }  ,fill opacity=0.23 ]   (338.89,11624.1) -- (348.57,11617.63) ;
			\draw [color={rgb, 255:red, 0; green, 0; blue, 0 }  ,draw opacity=1 ][fill={rgb, 255:red, 155; green, 155; blue, 155 }  ,fill opacity=0.23 ]   (327.57,11651.37) -- (344.64,11651.93) ;
			\draw [color={rgb, 255:red, 0; green, 0; blue, 0 }  ,draw opacity=1 ][fill={rgb, 255:red, 155; green, 155; blue, 155 }  ,fill opacity=0.23 ]   (289.69,11665.06) -- (306.22,11656.95) ;
			\draw    (56.96,11623.01) -- (71.35,11626.43) ;
			\draw    (53.48,11650.66) -- (67.23,11654.45) ;
			\draw    (71.35,11626.43) -- (77.56,11640.44) ;
			\draw    (67.23,11654.45) -- (77.56,11640.44) ;
			\draw    (47.27,11636.65) -- (56.96,11623.01) ;
			\draw    (47.27,11636.65) -- (53.48,11650.66) ;
			\draw    (56.96,11623.01) -- (67.23,11654.45) ;
			\draw    (56.96,11623.01) -- (77.56,11640.44) ;
			\draw    (71.35,11626.43) -- (53.48,11650.66) ;
			\draw    (71.35,11626.43) -- (67.23,11654.45) ;
			\draw    (71.35,11626.43) -- (47.27,11636.65) ;
			\draw    (47.27,11636.65) -- (77.56,11640.44) ;
			\draw    (47.27,11636.65) -- (67.23,11654.45) ;
			\draw    (99.63,11624.41) -- (114.3,11623.28) ;
			\draw    (103.94,11652.04) -- (118.09,11651.45) ;
			\draw    (114.3,11623.28) -- (124.08,11634.83) ;
			\draw    (118.09,11651.45) -- (124.08,11634.83) ;
			\draw    (94.15,11640.49) -- (99.63,11624.41) ;
			\draw    (94.15,11640.49) -- (103.94,11652.04) ;
			\draw    (99.63,11624.41) -- (118.09,11651.45) ;
			\draw    (99.63,11624.41) -- (124.08,11634.83) ;
			\draw    (114.3,11623.28) -- (103.94,11652.04) ;
			\draw    (114.3,11623.28) -- (118.09,11651.45) ;
			\draw    (114.3,11623.28) -- (94.15,11640.49) ;
			\draw    (94.15,11640.49) -- (124.08,11634.83) ;
			\draw    (94.15,11640.49) -- (118.09,11651.45) ;
			\draw    (77.56,11640.44) -- (94.09,11640.5) ;
			\draw  [fill={rgb, 255:red, 155; green, 155; blue, 155 }  ,fill opacity=0.23 ][dash pattern={on 0.84pt off 2.51pt}] (83.85,11660.35) .. controls (81.23,11660.8) and (78.49,11661.05) .. (75.67,11661.05) .. controls (57.26,11661.05) and (42.33,11650.73) .. (42.33,11637.99) .. controls (42.33,11625.26) and (57.26,11614.94) .. (75.67,11614.94) .. controls (78.47,11614.94) and (81.2,11615.18) .. (83.81,11615.63) -- cycle ;
			\draw    (56.7,11623.8) -- (53.7,11649.75) ;
			\draw    (99.97,11624.7) -- (103.62,11651.74) ;
			\draw  [fill={rgb, 255:red, 155; green, 155; blue, 155 }  ,fill opacity=0.23 ][dash pattern={on 0.84pt off 2.51pt}] (206.8,11604.28) .. controls (209.99,11606.48) and (211.06,11611) .. (209.19,11614.37) -- (183.51,11660.71) .. controls (181.64,11664.09) and (177.53,11665.04) .. (174.34,11662.84) -- (165.66,11656.86) .. controls (157.68,11651.36) and (155,11640.05) .. (159.68,11631.61) -- (175.19,11603.62) .. controls (179.87,11595.18) and (190.14,11592.8) .. (198.12,11598.3) -- cycle ;
			
			\draw (303.65,11674.93) node [anchor=north west][inner sep=0.75pt]    {$( c)$};
			\draw (189.56,11674.93) node [anchor=north west][inner sep=0.75pt]    {$( b)$};
			\draw (77.58,11675.11) node [anchor=north west][inner sep=0.75pt]    {$( a)$};

		\end{tikzpicture}
		\caption{{(a) Dumbbell graph $K_n\text{--}K_n$, (b) $d$-regular ring-of-cliques comprising of $m$ cliques connected in a ring, (c) Infinite $3$-regular tree. The shaded portions specify the Cheeger ``cut''.}
		}
		\label{fig:Isoperimetric_ratio_of_common_graphs}
	\end{figure}
	
	\begin{example}[``Good'', ``bad'' and ``ideal'' expanders]\label{ex:expandertypes}
		The Cheeger constant measures how robustly connected a graph is. For the complete graph $K_n$ on $n$ nodes, the Cheeger constant is $\lceil\tfrac{n}{2}\rceil$. For the path graph on $n$ nodes, which is a tree with two nodes of degree 1, and the other $n-2$ nodes of degree 2, the Cheeger constant is $\tfrac{2}{n}$. For the dumbbell graph $K_n\text{--}K_n$ comprising of two cliques $K_n$ joined by a bridge, the Cheeger constant is $O(\tfrac{1}{n})$; see Figure~\ref{fig:Isoperimetric_ratio_of_common_graphs}(a). 		Easily-disconnected graphs such as the path, dumbbell and $d$-regular ring-of-cliques comprising of $m$ cliques each of size $(d+1)$ connected in a ring (see Figure~\ref{fig:Isoperimetric_ratio_of_common_graphs}(b)) are bad expanders while well-connected graphs such as the complete graph are good expanders. The complete graphs form an expander family if we relax the requirement that the $d$-regular graph have a fixed $d$ independent of $n$. 	
		
		Intuitively, ideal graph expansion entails branching outward from each node, expending as few edges as possible on loops. This intuition is borne out by the $d$-regular infinite tree ($d\ge 3$), which is maximally expanding with isoperimetric ratio $h(G) = d-2$ \cite[\S 5]{hoory2006expander}; see Figure~\ref{fig:Isoperimetric_ratio_of_common_graphs}(c). On such a tree, the surface area to volume ratio of a sphere stays constant as the radius increases rather than tending to zero. In contrast, finite trees are bad expanders \cite{finitetreeLucaT}. A random $d$-regular graph for any fixed $d\ge 3$ and $n\gg d$ is almost an ideal expander  \cite{friedman2008proof}. A $\log_d n$-depth neighbourhood of such a graph is a $d$-regular tree with overwhelming probability \cite{bauerschmidt2017local}. Large sparse graph families cannot simultaneously have uniform expansion and nonnegative curvature  	\cite{salez2021sparse,munch2022mixing}.
	\end{example}
		
	The discrete Cheeger inequality \cite{cheeger1970lower,alon1984eigenvalues} shows that the spectral gap of $G$ provides an estimate of its Cheeger constant:
	\begin{align}\label{eq:expansionSpectralgap}
		\frac{d-\mu_2}{2} &\le h(G) \le \sqrt{2d(d-\mu_2)}.
	\end{align}
	In particular, the spectral gap is bounded away from zero if and only if $h(G)$ is bounded away from zero. A $d$-regular graph on $n$ nodes is a \emph{$(n,d,\alpha)$-spectral-expander} if $\tfrac{1}{d}\cdot {\mu}(A)\le \alpha$. From \eqref{eq:expansionSpectralgap}, we see that the geometric and spectral notions of expansion are intimately related. A small $\alpha$ (or large spectral gap) implies that random walks on the graph do not encounter any bottlenecks and mix rapidly (in $O(\log n)$ steps).

\subsection{Expander-based graph rewiring algorithms} \label{sec:RLEF-GRLEF}
In this section, we propose two graph rewiring algorithms motivated by an explicit expander construction and compare the evolution of graph expansion and curvature along the rewiring process of our algorithms with that of the Stochastic Discrete Ricci Flow (SDRF) algorithm introduced in \cite{topping2022understanding}. 

\subsubsection{The SDRF rewiring algorithm}  The SDRF algorithm \cite{topping2022understanding} is motivated from the idea that structural bottlenecks in the input graph, in the form of negatively curved edges, lead to information oversquashing. There, the notion of curvature, which can be construed as a ``local'' Cheeger constant, plays a central role in the rewiring process. Each iteration of the SDRF cycles between adding a ``supporting'' edge around a negatively curved edge and removing a ``redundant'' positively-curved edge with the goal of preserving the overall node degree distribution; see Figure~\ref{fig:RLEF_SDRF}(b). Since the addition and removal of edges are done independently of each other, the SDRF is \emph{non-local} and may disconnect a connected input graph. This issue is addressed in practice by way of a tunable hyperparameter that can be chosen so that no edges are removed \cite{topping2022understanding}.  Not removing any edges, however, may adversely impact the time and memory complexity of the downstream GNN as the rewired graph becomes denser. In the sequel, we shall only consider the ``default'' setting for SDRF where in each cycle, an edge is always removed so as to preserve the total edge count.

Compared to the SDRF, the rewiring algorithms we describe now aim at improving the ``global'' Cheeger constant $h(G)$ using only \emph{local} modifications without ever disconnecting the graph while preserving the degrees exactly; see Figure~\ref{fig:RLEF_SDRF}(a). 

\vspace{.1cm}
\subsubsection{The Random Local Edge Flip (RLEF) algorithm}\label{sec:RLEF}

\begin{figure}
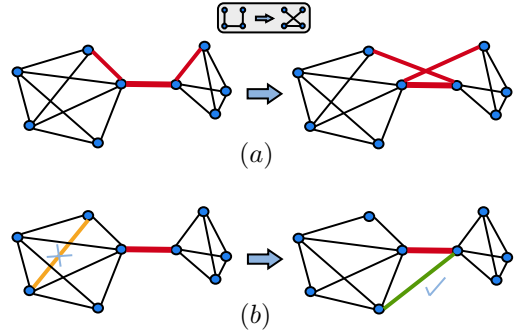

	\centering
	
	\tikzset{every picture/.style={line width=0.75pt}} 
	

	\caption{(a) One iteration of the Random Local Edge Flip (RLEF) primitive (shown in inset) on a dumbbell-like graph. Edge flips around the bridge edge alleviates the bottleneck while exactly preserving the degrees. (b) One iteration of the non-local Stochastic Discrete Ricci Flow (SDRF) transformation \cite{topping2022understanding}, which cycles between adding a ``supporting'' edge (in green) around the  bridge edge (in red) and removing a ``redundant'' edge (in orange). The addition and removal of edges are done independently of each other, and the red and orange edges can be {anywhere} in the graph. In this sense, unlike RLEF the SDRF transformation is \emph{non-local}  and may disconnect a connected graph.
	}
	\label{fig:RLEF_SDRF}
\end{figure}

The Random Local Edge Flip (RLEF) (Algorithm \ref{alg:RandomEdgeFlip}) is inspired from the {``Flip Markov chain''} \cite{mahlmann2005peer,feder2006local,allen2016expanders,cooper2019flip,expanderflipquasi} that transforms any regular connected graph into an expander with high probability. Given an input graph our goal is to construct an expander-like graph on the same set of nodes using a sequence of {local} transformations without disconnecting the graph or changing the degrees of the nodes. Starting from a connected graph $G^{(0)} = G= (\cV,\cE)$, the RLEF transformations define a Flip Markov chain, i.e., a sequence of connected graphs $G^{(1)},G^{(2)},\ldots,G^{(T)}$ over $\cV$, by iteratively sampling two distinct nodes $u, v \in \cV$ uniformly at random over the edges of $G^{(t)}$ and performing a ``local'' edge flip that amounts to exchanging a random neighbor; see inset in  Figure~\ref{fig:RLEF_SDRF}(a). The moves in this Flip Markov chain select a random three-path $i-u-v-j$ before executing the edge flip. This is done by first selecting a \emph{hub edge} $(u, v)$ uniformly at random and then selecting $i$ resp.\ $j$ as a random neighbor of $u$ resp.\ $v$, conditioned on the four nodes being distinct. By design, the RLEF transformation preserves the degree of all nodes as well as connectivity. For $d$-regular connected inputs, the Flip Markov chain converges to the uniform distribution over the set of connected $d$-regular graphs \cite{mahlmann2005peer,cooper2019flip}. For $d$-regular input $G$ on $n$ nodes, repeatedly applying the RLEF transformation produces a spectral expander in $O(d^2n^2 \sqrt{\log n})$ steps with high probability \cite{allen2016expanders}.

\begin{algorithm}
	\begin{algorithmic}[1]
		\Statex{\textbf{Input}: $G^{(t)}=(\cV,\cE^{(t)})$}
		\Statex{\textbf{Output}: $G^{(t+1)}=(\cV,\cE^{(t+1)})$}
		\State{Sample an edge $(u, v)\in \cE^{(t)}$ uniformly at random}
		\State{Sample a node $i$ uniformly at random from $N_{G^{(t)}}(u)$}
		\If{$i\in N_{G^{(t)}}(v)$ or $i=v$}
		\State{abort}
		\State \textbf{return} $G^{(t)}=(\cV,\cE^{(t)})$
		\Else
		\Repeat
		\State{Sample a node $j$ uniformly at random from $N_{G^{(t)}}(v)$}
		\Until{$j\not\in N_{G^{(t)}}(u)$ and $j\neq u$}
		\State{$\cE^{(t+1)} \gets \cE^{(t)}\setminus \{(i,u),(j,v)\}$}
		\State{$\cE^{(t+1)} \gets \cE^{(t)}\cup \{(i,v),(j,u)\}$}
		\State \textbf{return} $G^{(t+1)}=(\cV,\cE^{(t+1)})$
		\EndIf
	\end{algorithmic}
	\caption{\textbf{ 
			Random Local Edge Flip (RLEF)}}
	\label{alg:RandomEdgeFlip}
\end{algorithm}

As we empirically demonstrate in Section \ref{sec:experiments}, 
the evolution of the spectral expansion along the RLEF rewiring process has three distinct phases, namely, (a) an initial phase with essentially no expansion, (b) a relatively short intermediate phase characterized by rapid expansion, and (c) a stable phase when the expansion saturates; see Figure~\ref{fig:exp}. The slow initial rate of expansion of RLEF contrasts with that of the SDRF, which employs a greedy strategy by selecting the most negatively curved edges first \cite{topping2022understanding}. 
We now describe a greedy version of RLEF, the G-RLEF that allows us to accelerate the spectral expansion by sampling the hub edge $(u,v)$ non-uniformly in proportion to their ``effective resistance''.

\subsubsection{The Greedy Random Local Edge Flip (G-RLEF) algorithm}\label{sec:GRLEF}
For our greedy sampling strategy, we shall rely on physical metaphor viewing graphs as electrical networks \cite{doylesnellBook,lyonsPeresprobabilityTreesNwsBook}. For any two vertices $u$ and $v$ in the same connected component of $G$, the \emph{effective resistance} between $u$ and $v$ is defined as the energy dissipation when a unit of current is injected into one and a unit extracted at the other, and is given by the expression \cite{doylesnellBook}
\begin{align}\label{eq:EffectiveResistance}
	R_{uv} :=\left(\chi_{u}-\chi_{v}\right)^{T} L^{+}\left(\chi_{u}-\chi_{v}\right),
\end{align}
where $L^{+}$ is the Moore-Penrose pseudoinverse of the Laplacian $L$, and $\chi_u \in \Rb^{n\times 1}$ is the canonial basis vector with a 1 in position $u$. If the nodes $u,v$ reside in different connected components, then the associated effective resistance is infinite. 

Random spanning trees of a graph are intimately related to properties of electrical networks \cite{lyonsPeresprobabilityTreesNwsBook}. In particular, the effective resistance of an edge equals the probability that the edge appears in a uniformly random spanning tree of $G$ \cite{kirchhoff1847,chandra1996electrical}. The higher the effective resistance of an edge, the more likely it is for that edge to appear in a random spanning tree. Thus, intuitively, effective resistance captures the ``electrical importance'' of an edge. High resistance paths spanning bottlenecks (e.g., the bridge edge in the dumbbell-like graph in Figure~\ref{fig:RLEF_SDRF}) are more electrically important and such paths should be sampled with higher probability \cite{spielman2011graph}. 

Computing the effective resistance can be costly \cite{spielman2011graph}. Our greedy sampling scheme instead exploits a relationship between the effective resistance of an edge and its inverse triangle counts. For each pair $u,v \in \cV$, let $\sharp_{\Delta}(u, v) = |N_G(u)\cap N_G(v)|$ denote the number of common neighbors of $u$ and $v$. If $(u,v)$ is an edge, this coincides with the number of triangles that contain $(u,v)\in \cE$. Then the effective resistance $R_{uv}$ of any edge $(u,v) \in \cE$ satisfies the following inequality \cite{le2021edge,sotiropoulos2021triangle}:
\begin{align}\label{eq:effResTriangleUppBound}
	R_{uv} \le \frac{2}{2+\sharp_{\Delta}(u, v)}.
\end{align}

We now describe our greedy sampling strategy  (Algorithm~\ref{alg:G-RandomEdgeFlip}). G-RLEF first samples an edge $(u,v)$ according to their {inverse triangle count} (lines 1 and 2 in Algorithm~\ref{alg:G-RandomEdgeFlip}). We add an inverse temperature hyperparameter $\tau$ controlling the randomness of the sampling. Next, we choose the edges $(u, i)$, $(v, j)$ to be flipped in such a way that the net change in the number of triangles is as small as possible. This strategy is motivated by our observations in Section \ref{sec:experiments} (see Figure~\ref{fig:exp}) indicating that an increase in the spectral gap is concurrent with a decrease in the number of triangles. This observation can be explained as follows: Let $G$ be a $d$-regular graph. The number of triangles in $G$ is given by $\frac{1}{6}\text{Trace}(A(G)^3)$. Indeed, the entry $(i, i)$ of the adjacency matrix $A(G)$ counts the number of paths $i \to j \to k \to i$, which is twice the number of triangles that include $i$ as a node. Adjusting for overcounting, we obtain the above expression for the number of triangles in $G$. Hence the number of triangles converges to 0 as the spectral gap converges to 1. 

\begin{algorithm}
\begin{algorithmic}[1]
	\Statex{\textbf{Input}: $G^{(t)}=(\cV,\cE^{(t)})$, inverse temperature $\tau > 0$}
	\Statex{\textbf{Output}: $G^{(t+1)}=(\cV,\cE^{(t+1)})$}
	\State{Define $\pmb{x}_{(u, v)} := \frac{2}{2 + \sharp_{\Delta}(u, v)}$\Comment{$\sharp_{\Delta}(u, v)$ is the number of common neighbors of $u$ and $v$}}
	\State Sample edge $(u, v)$ with probability given by the $(u,v)$ component of $\text{softmax}(\tau \pmb{x})$.
	\If{$N_{G^{t}}(u) \subseteq N_{G^{t}}(v)$ or $N_{G^{t}}(v) \subseteq N_{G^{t}}(u)$}
	\State{abort}
	\State \textbf{return} $G^{(t)}=(\cV,\cE^{(t)})$
	\Else
	\State Choose $i \in N_{G^{t}}(u) \setminus N_{G^{t}}(v)$ such that $\sharp_{\Delta}(i, v) - \sharp_{\Delta}(i, u)$ is minimal
	\State Choose $j \in N_{G^{t}}(v) \setminus N_{G^{t}}(u)$ such that $\sharp_{\Delta}(j, u) - \sharp_{\Delta}(j, v)$ is minimal
	\State{$\cE^{(t+1)} \gets \cE^{(t)}\setminus \{(i,u),(j,v)\}$}
	\State{$\cE^{(t+1)} \gets \cE^{(t)}\cup \{(i,v),(j,u)\}$}
	\State \textbf{return} $G^{(t+1)}=(\cV,\cE^{(t+1)})$
	\EndIf
\end{algorithmic}
\caption{\textbf{ 
		Greedy Random Local Edge Flip (G-RLEF)}}
\label{alg:G-RandomEdgeFlip}
\end{algorithm}

\begin{remark}[Comparing G-RLEF and SDRF sampling] \label{rem:comparesamplers}%
G-RLEF samples edges according to inverse triangle counts while SDRF \cite{topping2022understanding} selects the most negatively curved edges first. The sampling mechanisms are similar in the sense that curvature on graphs is intrinsically related to the existence and relative abundance of triangles, which are the hallmarks of positive curvature \cite{jost2014ollivier}: The more the number of triangles two neighboring nodes share, the larger the overlap of their neighborhoods and hence larger the curvature of the edge between the two nodes. For a relation between different discrete notions of curvature and the effective resistance see, e.g., \cite{devriendt2022discrete}. Compared to the non-local SDRF, for the G-RLEF the repertoire of transformations is limited to the local edge flip operations. Locality comes with the advantage of never disconnecting the graph and preserving the degrees exactly. 
\end{remark}

\section{Experiments}\label{sec:experiments}
\begin{figure}
\centering
\includegraphics[scale=.2]{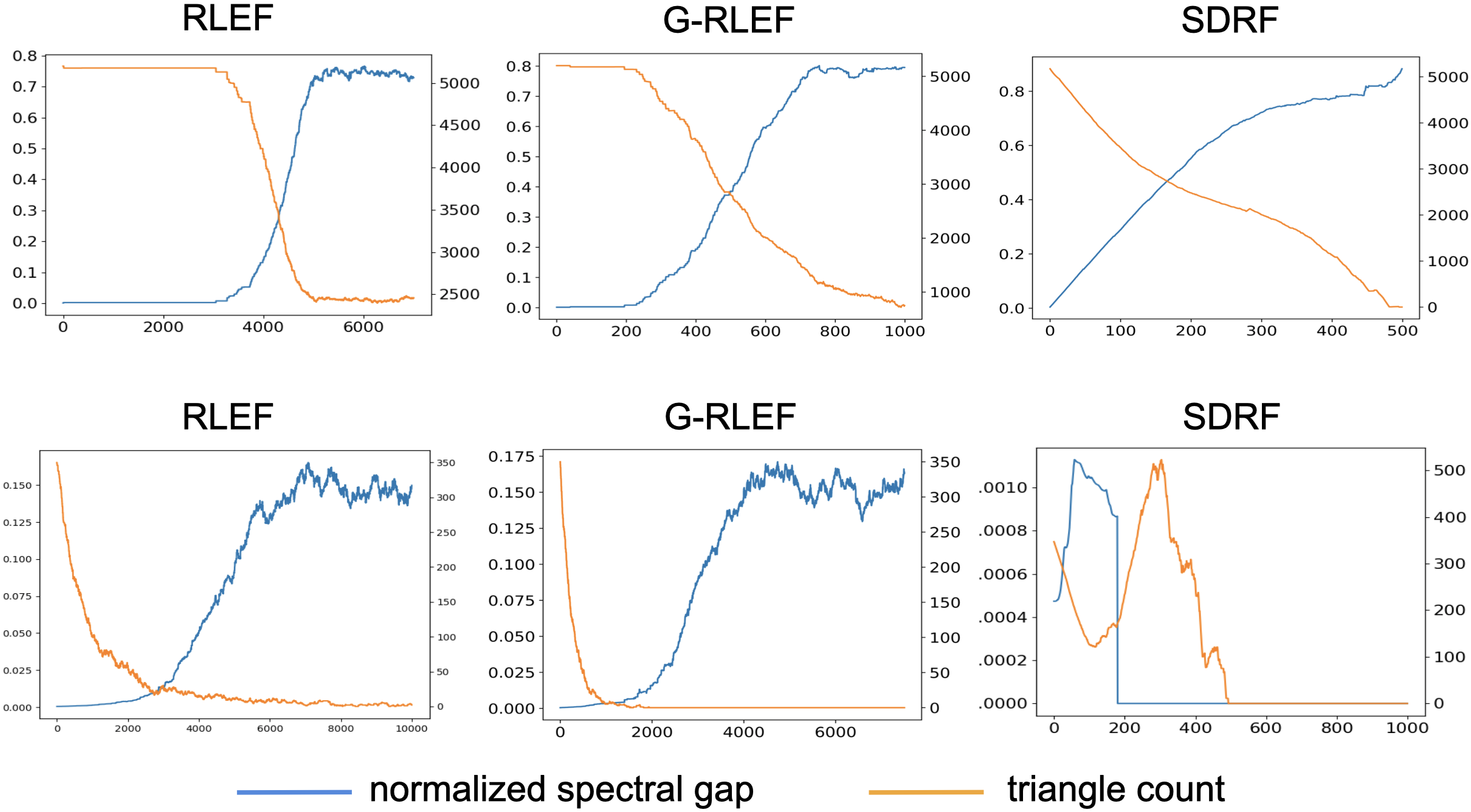}
\caption{Evolution of the normalized spectral gap (blue) and triangle counts (orange) for the RLEF, G-RLEF and SDRF as a function of the number of iterations for the dumbbell (top row), and the ring-of-cliques (bottom row). The faster rate of expansion for the non-local SDRF comes at the cost of disconnecting the input graph (bottom rightmost plot). 
}
\label{fig:exp}
\end{figure}

We empirically study the rewiring dynamics of our proposed algorithms and demonstrate their efficacy in alleviating oversquashing in a learning task modeled on the \textsc{NeighborsMatch} problem \cite{alon2021on}.
The code to reproduce our results is available at {\tt\small{\url{https://github.com/kedar2/Oversquashing}}}.
\paragraph{Rewiring dynamics for synthetic graphs}
We study the evolution of the spectral expansion and triangle counts for the RLEF, G-RLEF and SDRF along the rewiring process for two types of graphs:
(a) the dumbbell graph $K_n\text{--}K_n$ comprising of two cliques $K_n$ joined by a bridge, and (b) the $d$-regular ring-of-cliques, which consists of $m$ cliques each of size $(d+1)$ connected in a ring after removing one edge from every clique and connecting the endpoints of the removed edges \cite{mahlmann2005peer}; see Figures~\ref{fig:Isoperimetric_ratio_of_common_graphs}(a) and~\ref{fig:Isoperimetric_ratio_of_common_graphs}(b). Both the dumbbell and ring-of-cliques feature a low Cheeger constant and a high triangle density. In our experiments, we consider a dumbbell graph with $n = 50$ nodes, and a 4-regular ring-of-cliques with a total of $n = 250$ nodes. Figure~\ref{fig:exp} shows that the process of spectral expansion is closely related to the process of removal of triangles. This observation motivates G-RLEF's greedy sampling strategy and also explains its efficiency. For both the dumbbell and ring-of-cliques, G-RLEF consistently converges faster than RLEF. Between G-RLEF and SDRF, we see that the rate of spectral expansion is often faster for the SDRF as is the case for the dumbbell graph. This can be attributed to the fact that the SDRF employs a richer set of transformations (see Remark~\ref{rem:comparesamplers}).

The faster rate of expansion for the SDRF often comes at the cost of consistency: Both RLEF and G-RLEF are local algorithms -- they add and remove edges at the same location, preserving the node degrees and connectivity of the graph. These guarantees do not exist for the non-local SDRF: For the ring-of-cliques, the number of triangles as well as the spectral gap fluctuate unpredictably until the graph becomes disconnected; see Figure~\ref{fig:exp}.
Recall that the SDRF operates by ``supporting'' a negatively curved edge at a bottleneck with more edges around it. Since these new supporting edges will also be located around a bottleneck, they are more likely to be negatively curved. For the dumbbell, the overall curvature (as roughly measured by triangle counts) along the SDRF rewiring process decreases despite increasing the curvature locally around the bottleneck edge. This effect is caused by the {volume} of negatively curved edges increasing rather than any particular edge becoming more negatively curved. 

\begin{figure}
\centering
\includegraphics[scale=.18]{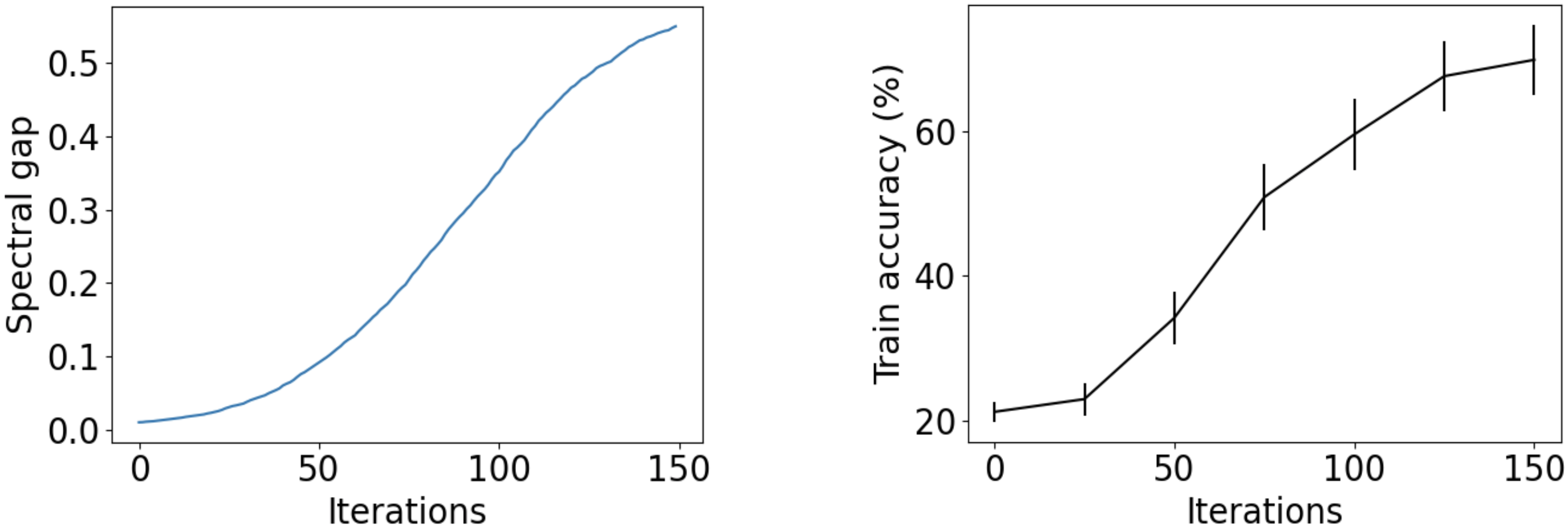}
\caption{
Evolution of the normalized spectral gap and training accuracy as a function of the number of G-RLEF rewiring iterations for a learning task modeled on the \textsc{NeighborsMatch} problem for a path-of-cliques input.
}
\label{fig:exp2}
\end{figure}

\paragraph{The \textsc{NeighborsMatch} problem}
We tested the efficacy of G-RLEF in alleviating oversquashing in a learning task modeled on the \textsc{NeighborsMatch} problem \cite{alon2021on} described in Section~\ref{sec:neighborsmatch}. Given an input graph $G$, a target node $\textsf{T}$, and a subset $\mathcal{S}$ of the nodes of $G$, we assign each node in $\mathcal{S}$ a different random $|\mathcal{S}|$-dimensional one-hot vector encoding the number of orange neighbors. Likewise, we represent the label of $\textsf{T}$ as a random $|\mathcal{S}|$-dimensional one-hot vector and the goal is to predict the node $\textsf{T'} \in \mathcal{S}$ with the same label as $\textsf{T}$. We take the input $G$ to be a {path-of-cliques}, which comprises of three cliques each of size 10 connected in a path by two edges, one edge between the last node of clique-1 and the first node of clique-2, and one edge between the last node of clique-2 and the first node of clique-3. The set $\mathcal{S}$ consists of the first 9 nodes in clique-1, and the target $\textsf{T}$ is the last node of clique-3. The problem radius, i.e., the maximum distance between $\textsf{T}$ and a node in $\mathcal{S}$ is $r=5$. We trained a graph attention network \cite{velivckovic2017graph} with $r+1$ layers each of width 64 on a training dataset comprising of 10000 copies of $G$, where each copy has a different mapping matching the target as described above. Figure~\ref{fig:exp2} shows the evolution of the normalized spectral gap and training accuracy as a function of the number of G-RLEF iterations. We observe that both the normalized spectral gap and training accuracy increase monotonically in the number of G-RLEF iterations until they saturate at around $150$ iterations.

\section{Conclusion}
We presented a framework for analyzing the information decay arising from oversquashing in GNNs guided by a model of noisy computation due to von Neumann. We proposed a local graph rewiring algorithm G-RLEF motivated by an expander graph construction employing an effective resistance based edge sampling strategy. We compared its spectral expansion and curvature properties with that of an existing non-local rewiring strategy SDRF \cite{topping2022understanding}. Insofar as the expansion properties of the input graph are an important determinant of oversquashing in GNNs, our rewiring algorithm offers potential advantages over existing approaches in terms of locality, preservation of graph connectivity and node degrees. In future work, it remains to test the efficacy of our algorithm in alleviating oversquashing on long-range graph learning benchmark datasets.

\section*{Acknowledgment}
The authors thank Anuran Makur, Chuteng Zhou, Florentin M\"unch and J\"urgen Jost for helpful discussions. This project has received funding from the European Research Council (ERC) under the EU's Horizon 2020 research and innovation programme (grant agreement n\textsuperscript{o} 757983).

\bibliographystyle{IEEEtran}
\bibliography{IEEEabrv,oversquashing}

\end{document}